\documentclass{article}

\usepackage[preprint]{neurips_2022}

\usepackage[utf8]{inputenc} 
\usepackage[T1]{fontenc}    
\usepackage{hyperref}       
\usepackage{url}            
\usepackage{booktabs}       
\usepackage{amsfonts}       
\usepackage{nicefrac}       
\usepackage{microtype}      
\usepackage{xcolor}         
\usepackage{graphicx}
\usepackage{amsmath}
\usepackage{algorithm}
\usepackage{algorithmic}
\usepackage{array}

\title {Matrix-Transformation Based Low-Rank Adaptation (MTLoRA): A Brain-Inspired Method for Parameter-Efficient Fine-Tuning}

\author{%
  Yao Liang$^{1,2,\ddagger}$, \quad
  Yuwei Wang$^{1,\ddagger}$, \quad
  Yang Li$^{1,2}$, \quad
  Yi Zeng$^{1,2,3}$\thanks{Corresponding authors} \\
  $^{1}$Brain-inspired Cognitive Intelligence Lab, Institute of Automation, \\Chinese Academy of Sciences, Beijing, China\\
  $^{2}$School of Artificial Intelligence, University of Chinese Academy of Sciences, Beijing, China\\
  $^{3}$Key Laboratory of Brain Cognition and Brain-inspired Intelligence Technology, \\Chinese Academy of Sciences, Shanghai, China\\
  $^{\ddagger}$Co-first authors with equal contribution\\
  \texttt{\small \{liangyao2023,yuwei.wang,liyang2019,yi.zeng\}@ia.ac.cn}
}

\begin{document}

\maketitle

\begin{abstract}
Fine-tuning techniques based on Large Pretrained Language Models (LPLMs) have been proven to significantly enhance model performance on a variety of downstream tasks and effectively control the output behaviors of LPLMs. Recent studies have proposed numerous methods for fine-tuning a small number of parameters based on open-source LPLMs, reducing the demand for computational and storage resources. Among these, reparameterization fine-tuning methods represented by LoRA (Low-Rank Adaptation) have gained popularity. We find that although these methods perform well in many aspects, there is still considerable room for improvement in terms of complex task adaptability, performance, stability, and algorithm complexity. 
In response to this, inspired by the idea that the functions of the brain are shaped by its geometric structure, this paper integrates this idea into LoRA technology and proposes a new matrix transformation-based reparameterization method for efficient fine-tuning, named Matrix-Transformation based Low-Rank Adaptation (MTLoRA). The spatiotemporal patterns of brain neural activity are the excitation of different wavelength characteristic patterns of its geometric structure. MTLoRA aims to dynamically alter its spatial geometric structure by applying a transformation-matrix $T$  to perform linear transformations, such as rotation, scaling, and translation, on the task-specific parameter matrix, generating new matrix feature patterns (eigenvectors) to mimic the fundamental influence of complex geometric structure feature patterns in the brain on functions, thereby enhancing the model's performance in downstream tasks. The transformation-matrix $T$  contains four different structures, each designed to simulate the geometric feature patterns of the brain at different levels. In Natural Language Understanding (NLU) tasks, it is evaluated using the GLUE benchmark test, and the results reveal that MTLoRA achieves an overall performance increase of about 1.0\% across eight tasks and reduces the standard deviation by 0.7\% in the Corpus of Linguistic Acceptability (CoLA) task; in Natural Language Generation (NLG) tasks, MTLoRA improves performance by an average of 0.95\% and 0.56\% in the DART and WebNLG tasks, respectively.
\end{abstract}

\section{Introduction}

In recent years, with the rapid development of large pre-trained language models (LPLMs) such as BERT~\citep{devlin2019bert} and GPT-3~\citep{brown2020language}, these models have shown exceptional performance in many downstream tasks of natural language processing (NLP), including text generation, machine translation, sentiment analysis, and question-answering systems~\citep{radford2019language,he2020deberta,raffel2020exploring,devlin2019bert,liu2019roberta,peters-etal-2018-deep,brown2020language}. However, training and deploying LPLMs from scratch face significant resource challenges, as training a GPT-3 model with 175 billion parameters requires running hundreds of NVIDIA A100 40GB GPUs for approximately 200 days. Such a level of resource investment is unaffordable for most research institutions and enterprises.

Fine-tuning techniques based on LPLMs have been proven to effectively enhance model performance~\citep{ouyang2022training,wei2021finetuned,min2021metaicl,wang2022self,liu2022few}, enabling the model to acquire desired capabilities and discard unwanted ones~\citep{ouyang2022training,askell2021general}. The rise of open-source LPLMs globally, such as GPT~\citep{radford2019language}, BERT~\citep{devlin2019bert}, GLM~\citep{du2022glm,zeng2022glm}, LLaMA~\citep{touvron2023llama1}, and LLaMA2~\citep{touvron2023llama2}, offers new opportunities for a wide range of research organizations and enterprises. These entities can use these open-source LPLMs as a foundation, combined with their industry knowledge and data, for full-parameter fine-tuning (Full Fine-Tuning)~\citep{qiu2020pre,raffel2020exploring}. However, even full-parameter fine-tuning based on open-source LPLMs may require significant computational and storage resources. For instance, according to a study by~\cite{dettmers2024qlora}, full-parameter fine-tuning of the LLaMA 65B model could require up to 20 NVIDIA A100 40GB GPUs, which remains prohibitively expensive for many organizations. Moreover, considering the diversity of application scenarios, each downstream application needs to be trained separately, occupying the same space as the original LPLM, which not only increases deployment costs but also challenges the reliability and stability of online services.

Recent research advancements have proposed many parameter-efficient fine-tuning methods based on LPLMs, aimed at reducing the demand for GPU computational power and storage resources during the training of downstream tasks. In the full-parameter fine-tuning process, all trainable parameters of the LPLM need to be updated. However, during the training process of parameter-efficient fine-tuning methods, the original parameters of the LPLM are frozen, and only a small number of trainable parameters are updated through gradients. This change is comparable in performance to full-parameter fine-tuning and significantly reduces the demand for computational power and storage.

In the field of parameter-efficient fine-tuning technology research for LPLMs, there are mainly three types of methods: First, Addition-based methods insert small trainable extension structures into the layers of the LPLM. During fine-tuning, only the trainable parameters within the extension structures are updated, while the original model's structure and parameters remain unchanged. A representative work in this category is Adapter Tuning~\citep{houlsby2019parameter,rebuffi2017learning,he2022towards}, which may increase the model's depth and complexity. Second, Specification-based methods designate certain parameters within the LPLM as trainable and freeze the rest. BitFit, for instance, updates only the model's bias parameters during fine-tuning~\citep{zaken2021bitfit}. Lastly, Reparameterization-based methods transform existing parameters into trainable ones. A pivotal work is LoRA (Low-Rank Adaptation), proposed by~\cite{hu2022lora}. LoRA approximates the increment matrix $\Delta W$ of the pre-trained model parameters $W_0$ through the product of low-rank decomposition matrices $A$ and $B$. Since the low-rank dimension $r$ of $A$ and $B$ is much smaller than the dimension of $W_0$, and the parameter matrix $W_0$ is frozen during training, updating only the $A$ and $B$ parameter matrices significantly reduces the amount of trainable parameters. For instance, fine-tuning the GPT-3 175B model with LoRA can reduce the number of training parameters by a factor of ten thousand and the GPU memory requirement by a factor of three. There are mainly two types of methods based on LoRA improvements: one is performance optimization, such as AdaLoRA~\citep{zhang2023adaptive}, IncreLoRA~\citep{Zhang2023IncreLoRAIP}, and DELTA-LoRA~\citep{zi2023delta}, which optimize LoRA in various aspects to enhance model performance. The other is functional expansion, like LongLoRA~\citep{chen2023longlora} and QLoRA ~\citep{dettmers2024qlora}, which expand the model's applicability in long contexts and quantization based on LoRA as a foundational component. The LoRA fine-tuning technique, with its advantages of excellent performance, simple structure, efficient training, and no inference delay, is expected to continue to drive the development and industrial application of large models.

Despite significant progress in fine-tuning with reparameterization methods~\citep{hu2022lora,chen2023longlora,zhang2023adaptive,zi2023delta} represented by LoRA, it still faces several key challenges: First, the structure of the parameter decomposition matrix is too simple and singular, making it difficult to dynamically represent various semantically complex downstream tasks. Second, performance fluctuation is significant. According to experiments by~\cite{hu2022lora}, LoRA has high performance fluctuation in the Corpus of Linguistic Acceptability (CoLA) task, with a standard deviation reaching 1.2\%. Lastly, there is the issue of computational complexity. AdaLoRA~\citep{zhang2023adaptive} and IncreLoRA~\citep{Zhang2023IncreLoRAIP} optimize the size of the parameter incremental matrix rank ($r$) through a mechanism that measures the importance scores of the singular value decomposition triplets of the parameter incremental matrix, allowing for different amounts of trainable parameters in incremental matrices at different positions. However, this requires iterative calculations of the importance scores of the triplets in more parameter decomposition matrices to dynamically adjust the size of the incremental matrix rank ($r$), significantly increasing the computational complexity. These challenges limit the application of LPLMs in a broader domain, and addressing these issues is crucial for advancing the development of large language models.

This study is inspired by the idea that the brain's functionality is shaped by its geometric structure~\citep{pang2023geometric} and integrates this concept into LoRA technology. We propose a novel matrix transformation-based reparameterization method for efficient fine-tuning named MTLoRA. The spatiotemporal patterns of brain neural activity are the excitation of different wavelength characteristic modes of its geometric structure~\citep{pang2023geometric}. MTLoRA employs a transformation-matrix $T$ to perform linear transformations on task-specific parameter matrices, such as rotation, scaling, and translation, dynamically altering their spatial geometric structure to generate new matrix feature patterns (eigenvectors). This mimics the fundamental impact of complex geometric structure characteristic patterns in the brain on functionality, enhancing the performance of the fine-tuned model, thereby alleviating the aforementioned issues. The $T$ transformation matrix contains four different structures, each designed to simulate the characteristic patterns of geometric structures at different levels of the brain.

In this paper, through empirical experiments on two categories of 11 tasks in NLU and NLG, the effectiveness of the MTLoRA method is verified. Compared to the LoRA method, it is evident from the experimental results table \ref{Table1} that MTLoRA improves performance by about 1.54\% ($\sigma$=0.1\%) on the CoLA~\citep{warstadt2019neural} task, effectively reducing the standard deviation; it enhances performance by about 3.61\% ($\sigma$=0.8\%) on the RTE~\citep{dagan2005pascal,haim2006second,giampiccolo2007third,bentivogli2009fifth} task; it increases performance by about 2.45\% ($\sigma$=0.3\%) on the MRPC~\citep{dolan2005automatically} task; and it boosts performance by about 0.88\% ($\sigma$=0.0\%) on the QQP (Quora Question Pairs) task. From the experimental results table \ref{Table2}, it is known that MTLoRA achieves an average performance increase of about 0.95\% ($\sigma$=0.1\%) on the DART~\citep{nan2020dart} task and about 0.56\% ($\sigma$=0.1\%) on the WebNLG~\citep{gardent2017webnlg} task.

Overall, the MTLoRA fine-tuning method proposed in this paper demonstrates significant advantages in downstream tasks based on LPLMs, which include:

1) The rich matrix transformation structure significantly enhances model performance in many tasks while maintaining the simplicity of the algorithm;

2) Effectively reduces performance fluctuations in certain tasks;

3) After training is completed, the incremental parameter matrix $\Delta W$ can be merged with the original parameter matrix $W_0$, achieving no additional latency during the inference stage;

4) Compared to full fine-tuning, the MTLoRA method significantly increases training speed, drastically reduces the amount of trainable parameters by 99\%, and lowers the hardware requirements by several times.

\begin{figure}[htb]
  \centering
  \includegraphics [width=14cm]{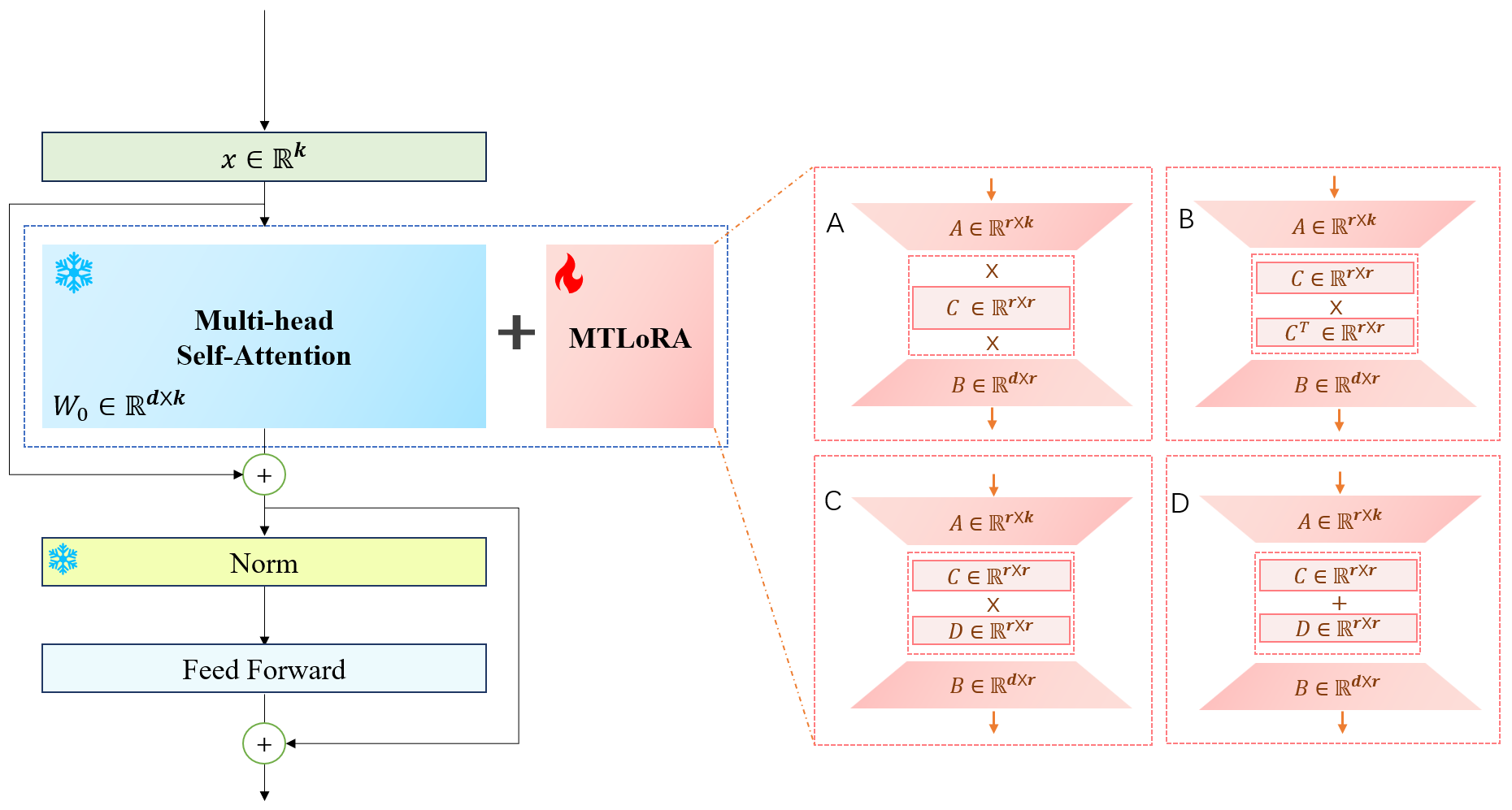}
  \caption{Structure of the MTLoRA fine-tuning method.}
  \label{P1}
\end{figure}

\section{Related work}

In recent years, large language models based on the Transformer~\citep{vaswani2017attention} architecture have made significant advancements in NLP field. Among these, BERT (Bidirectional Encoder Representations from Transformers) and its derivatives such as RoBERTa (Robustly Optimized BERT Approach) and DeBERTa (Decoding-enhanced BERT with Disentangled Attention) primarily utilize the Encoder part of the Transformer architecture~\citep{devlin2019bert,liu2019roberta,he2020deberta}. In contrast, the GPT (Generative Pre-trained Transformer) series, including GPT-2 and GPT-3, mainly rely on the Decoder part~\citep{radford2019language,brown2020language}. The core feature of these models is the Multi-Head Self-Attention mechanism, which can effectively capture the dependencies between different positions in a sequence. The basic algorithm of this mechanism can be represented by the following formula~\citep{vaswani2017attention}:
\begin{equation}\label{eq1}
\begin{aligned}
\text{MultiHead}(Q, K, V) &= \text{Concat}(head_1, \ldots, head_h)W^O  \\
head_i &= \text{Attention}(QW_i^Q, KW_i^K, VW_i^V)  \\
\text{Attention}(Q, K, V) &= \text{softmax}\left(\frac{QK^T}{\sqrt{d_k}}\right)V
\end{aligned}
\end{equation}
where \(Q\), \(K\), and \(V\) represent the Query, Key, and Value vectors, respectively; \(W_i^Q\), \(W_i^K\), and \(W_i^V\) are the corresponding linear transformation matrices; \(W^O\) is the output transformation matrix; and \(d_k\) is the dimension of the key vector, used to scale the dot product to prevent the softmax function from entering the saturation region.

The relationship between model performance and the amount of parameters has attracted the attention of numerous scholars. For instance, the GPT-2 model presented by~\cite{radford2019language}, which has 1.5 billion parameters, demonstrated a significant improvement in the quality and consistency of language generation compared to its predecessor, GPT. Furthermore, the GPT-3 model introduced by Brown et al. in 2020, with its 175 billion parameters, exhibited unprecedented language understanding and generation capabilities~\citep{brown2020language}.

\textbf{Fine-tuning} large-scale models is a crucial process that enables pre-trained language models to adapt to specific downstream tasks, thereby playing a more significant role across various application scenarios~\citep{ouyang2022training,wei2021finetuned,min2021metaicl,wang2022self,liu2022few}. Although traditional fine-tuning methods have achieved certain successes in performance, they usually require updating a large number of parameters, which not only increases computational and storage burdens but may also lead to overfitting, especially in scenarios with scarce data~\citep{qiu2020pre,raffel2020exploring}. Against this backdrop, the emergence of parameter-efficient fine-tuning methods provides a new direction for model fine-tuning by updating a small number of trainable parameters, achieving more efficient and flexible fine-tuning. 

\textbf{LoRA (Low-Rank Adaptation)} is a novel reparameterization fine-tuning technique proposed by~\cite{hu2022lora}. It is designed to facilitate fine-tuning training for adapting large pre-trained language models to downstream tasks with lower computational and storage resources. LoRA approximates the incremental matrix \(\Delta W\) of the pre-trained model parameter matrix \(W_0\) by the product of two low-rank parameter matrices \(B\) and \(A\), where \(B \in \mathbb{R}^{d \times r}\) and \(A \in \mathbb{R}^{r \times k}\), and the rank (\(r\)) is much smaller than the dimensions \(d\) or \(k\) in \(W_0 \in \mathbb{R}^{d \times k}\). During gradient updates, only the \(A\) and \(B\) parameter matrices are updated, with the \(W_0\) matrix being frozen, significantly reducing the number of trainable parameters and improving the performance of fine-tuning training. Specifically, the mathematical expression of LoRA during forward propagation is:
\begin{equation}\label{eq2}
\begin{aligned}
h = W_0 x + \Delta W x = W_0 x + BA x
\end{aligned}
\end{equation}
The \(A\) matrix is initialized with random Gaussian values, while the \(B\) matrix is initialized as a zero matrix. This initialization strategy is very important for maintaining the stability of the model at the beginning of training.

\textbf{AdaLoRA}, a fine-tuning method based on LoRA, was proposed by~\cite{zhang2023adaptive}. This method aims to find the optimal rank (\(r\)) size of the incremental parameter matrices at different positions in LPLMs to improve performance and prevent overfitting. The study found that parameter matrices at different positions, such as the word embedding projection matrix (\(W_e\)), the query/key/value projection matrices (\(W_q\), \(W_k\), \(W_v\)), and the output projection matrix (\(W_o\)) as well as the two weight matrices in the two layers of FFNs (\(W_{f1}\), \(W_{f2}\)), have different impacts on model performance across various tasks.

AdaLoRA approximates the incremental matrix \(\Delta W\) of the pre-trained model parameter matrix using singular value decomposition \(P \Lambda Q\), where \(\Lambda \in \mathbb{R}^r\) is a diagonal matrix containing \(r\) singular values represented by a one-dimensional vector, \(P \in \mathbb{R}^{d_1 \times r}\), and \(Q \in \mathbb{R}^{r \times d_2}\). To improve performance, AdaLoRA introduced a metric strategy based on the importance scores of the singular value decomposition triplets (left singular vector, singular value, right singular vector), allowing important triplets with high scores to be retained and unimportant triplets with scores below a threshold to be discarded, thereby effectively controlling the amount of trainable parameters in different incremental matrices and preventing overfitting. They found that for the NLP task SQuAD 2.0, AdaLoRA could improve the model's F1 score by 1.2\%. However, AdaLoRA generally requires finding the appropriate rank (\(r\)) in all parameter matrices, which brings higher computational costs.

\textbf{DELTA-LoRA}, proposed by~\cite{zi2023delta}, is an improved technique based on the LoRA. This technique enhances the model's expressive capability by updating more model parameters, effectively adapting to and handling complex downstream tasks. Compared to LoRA, DELTA-LoRA mainly differs in two aspects:
firstly, during the training process, in addition to updating the low-rank adaptation parameter matrices \(B\) and \(A\), it is also necessary to update the pre-trained model parameters \(W\), with the update expression for \(W\) as follows:
\begin{equation}\label{eq3}
\begin{aligned}
W^{(t+1)} = W^{(t)} + \lambda \cdot \alpha / r \cdot (A^{(t+1)} B^{(t+1)} - A^{(t)} B^{(t)})
\end{aligned}
\end{equation}
where the hyperparameters \(\lambda\), \(\alpha\), and \(r\) represent the update rate, scaling factor, and rank size, respectively.
Additionally, DELTA-LoRA removes the Dropout layer.

\textbf{QLoRA} is a quantization fine-tuning method based on LoRA, proposed by~\cite{dettmers2024qlora}. This method focuses on how to effectively reduce memory usage during the training phase of LPLMs while maintaining model performance. Unlike traditional quantization techniques applied only at the model deployment and inference stage, QLoRA proposes a new solution for memory optimization during the model training phase.

Quantization techniques typically reduce memory usage by lowering the numerical precision of the model, but this may sacrifice model performance. However, QLoRA, through NormalFloat, Double Quantization, and Paged Optimizers techniques, combined with the LoRA fine-tuning method, effectively resolves the contradiction between memory saving and performance maintenance. Specifically, first, the parameter matrices in LPLMs are quantized to 4-bit representations through NormalFloat, then, the quantization constants are further reduced through Double Quantization, reducing memory overhead. Paged Optimizers address potential memory shortages during GPU operation using NVIDIA's unified memory feature.

According to~\cite{dettmers2024qlora}, full fine-tuning a 65B model requires 780GB of GPU memory, while using QLoRA technology can reduce the GPU memory capacity requirement to 48GB per card, without sacrificing runtime and predictive performance.

\textbf{LongLoRA}, proposed by~\cite{chen2023longlora}, is an extension technique based on the LoRA fine-tuning method, aimed at optimizing existing large language models to handle longer contexts. The LongLoRA method has two main innovations: (1) applying the LoRA method to the Embedding layer and Normalization layer; (2) proposing a new shifted short attention mechanism. LongLoRA significantly extends the model's long context processing capability.

According to~\cite{chen2023longlora}, after applying LongLoRA fine-tuning to the LLaMA2-7B model~\citep{touvron2023llama2}, the model's context processing length could be expanded from 4096 to 100k. And after fine-tuning the LLaMA2-70B model on 8 NVIDIA A100 GPUs, its context length capability could be enhanced to 32k.

\section{The Method}

Our proposed MTLoRA fine-tuning approach is inspired by the idea that the brain's functionality is shaped by its geometric structure~\citep{pang2023geometric}. Specifically, the neural activities of the brain are incited by the inherent resonance modes of its geometric topological structure, and the task-induced activations are the excitation of whole-brain modes, with wavelengths exceeding 60mm. Furthermore, the dominant role of wave-like activity explains the close connection between the brain's geometric structure and its function, and wave dynamics are utilized to reconstruct the spatiotemporal characteristics of spontaneous and task-induced brain activity recordings.

The intrinsic resonance modes of the geometric structure of the brain's neocortex can be characterized by its geometric feature patterns~\citep{melrose1991electromagnetic,nozari2020brain}. These geometric feature patterns can be obtained by solving the eigenvalue problem of the Laplace–Beltrami Operator (LBO)~\citep{chavel1984eigenvalues,seo2011laplace} constructed on the basis of the group average template mesh representation~\citep{fischl1999high} of the neocortex. The geometric feature patterns obtained in this manner include physical properties such as the curvature of the neocortex surface and the spatial relationships between vertices in the mesh~\citep{wachinger2015brainprint}, analogous to how the resonance frequencies of a violin string are determined by the string's density, tension, and length. In essence, geometric feature patterns represent the vibrational modes of the system's dynamics~\citep{levy2006laplace}, and the spatiotemporal patterns of neural activity in the neocortex are the excitations of these geometric feature patterns, just as the harmonics produced by plucking a violin string are vibrations of its own resonance modes.

The spatiotemporal patterns of neural activity in the neocortex can be decomposed into a weighted sum of geometric feature patterns of different wavelengths~\citep{nowack1995neocortical,robinson2016eigenmodes}. According to the experimental results of ~\citep{pang2023geometric}, this decomposition can reconstruct the functional magnetic resonance imaging (fMRI) experimental data of neocortical activity obtained under spontaneous and task-induced conditions with more than 80\% accuracy, thereby confirming that the brain's geometric structure shapes its function. This relationship between the brain's geometric structure and function can also be extended to subcortical activities, such as those in the thalamus, striatum, and hippocampus, indicating that the close connection between geometric structure and function is a ubiquitous presence in the brain.

We believe LPLMs should also possess rich and complex spatial geometric structure feature patterns. We aim to find a fine-tuning method that can flexibly change the geometric structure of parameter matrices based on the characteristics of downstream tasks, generating new matrix feature patterns (eigenvectors), to simulate the impact of complex geometric structure feature patterns in the brain on function, thereby improving model performance.

This section will introduce MTLoRA fine-tuning method in detail, using large pre-trained language models based on the Transformer architecture as an example. Transformer-based models, such as the RoBERTa model, contain multiple Blocks, each with numerous dense parameter matrices, including word embedding projection matrices (\(W_e\)), query/key/value projection matrices (\(W_q\), \(W_k\), \(W_v\)), intermediate layer projection matrices (\(W_m\)), output layer projection matrices (\(W_o\)), and weight matrices in the MLP layer (\(W_f\)), among others. These parameter matrices carry rich semantic information, which is our optimization target.

MTLoRA aims to use a transformation-matrix $T$ to perform linear transformations on parameter matrices specific to a task, changing their spatial geometric structure, and generating new matrix feature patterns, to mimic the fundamental impact of complex geometric structure feature patterns in the brain on function, thereby improving the model's performance after fine-tuning training. Specifically, MTLoRA uses the product of low-rank parameter increment matrices \(A \in \mathbb{R}^{r \times k}\), \(B \in \mathbb{R}^{d \times r}\), and \(T \in \mathbb{R}^{r \times r}\), i.e., \(B(T)A\), to approximate the increment matrix \(\Delta W \in \mathbb{R}^{d \times k}\) of the pre-trained parameter matrix \(W_0 \in \mathbb{R}^{d \times k}\) in the LPLM, where rank \(r\) << \(k\) or \(d\). During training, \(W_0\) is frozen, and $T$ performs task-specific adaptive linear transformations on \(A\) and \(B\), including scaling, rotation, translation, reflection, skewing, and projection. $T$ contains four different structures, each designed to simulate different levels of geometric feature patterns in the brain, where structure 1 is the most basic and can handle most downstream application scenarios, while the other structures are more inclined to handle specific application scenarios with scarce or abundant corpora:

\textbf{Structure 1 SHIM (Spatial Harmonic Integration Matrix)}: This structure is designed to integrate different spatial feature transformations, similar to how various frequencies of harmonics combine in music to produce rich and complex timbres. By introducing the transformation matrix \(C\), it applies linear transformations such as spatial rotation, scaling, translation, and shearing to the \(A\) and \(B\) parameter matrices based on specific tasks, forming a comprehensive feature representation. This integration not only enhances the model's adaptability to different tasks but also provides a richer expression capability for adjusting model parameters, enabling the model to better adapt and optimize the processing effects of specific tasks. Its forward propagation mathematical expression in the model is shown as \ref{eq4}:
\begin{equation}\label{eq4}
\begin{aligned}
h = (W_0 + \Delta W) x = W_0 x + \Delta W x = W_0 x + (BCA) x
\end{aligned}
\end{equation}
where \(C \in \mathbb{R}^{r \times r}\). \(x\) represents the input vector, and $h$ represents the output vector. Because $r$ << $k$ or $d$, the number of parameters in the \(C\) transformation matrix is very small, adding minimal overhead. The \(A\) and \(C\) matrices are initialized using random Gaussian initialization, while the \(B\) matrix is initialized as a zero matrix, as illustrated in Figure \ref{P1} (A).

\textbf{Structure 2 ICFM (Intrinsic Correlation Feature Matrix)}: This structure, by introducing the transformation matrix \(C\) and its transpose \(C^T\), results in a positive semi-definite matrix after multiplication, which can simulate a covariance matrix capturing the intrinsic correlation structure within the feature space. This positive semi-definite matrix applies linear transformations such as rotation and scaling to the \(A\) and \(B\) parameter matrices, highlighting task-specific features within the feature space, thus enhancing the model's representational ability by increasing the differentiation between patterns. This structure shares similarities with the covariance matrix in Mahalanobis distance and spatial rotation in Principal Component Analysis (PCA)~\citep{wold1987principal}. Its forward propagation mathematical expression in the model is shown as \ref{eq5}:
\begin{equation}\label{eq5}
\begin{aligned}
h = (W_0 + \Delta W) x = W_0 x + \Delta W x = W_0 x + B(CC^T)A x
\end{aligned}
\end{equation}
where \(C \in \mathbb{R}^{r \times r}\), \(C^T \in \mathbb{R}^{r \times r}\), representing the transpose of matrix \(C\), \(C^{\prime} \in \mathbb{R}^{r \times r}\), as illustrated in Figure \ref{P1} (B).

\textbf{Structure 3 CTCM (Composite Transformation Coupling Matrix)}: By introducing two different matrices, \(C\) and \(D\), and forming a composite transformation matrix through matrix multiplication, this structure enhances the expressiveness of the parameter increment matrix, thereby fostering the model's adaptability to complex tasks. Its forward propagation mathematical expression in the model is shown as \ref{eq6}:
\begin{equation}\label{eq6}
\begin{aligned}
h = (W_0 + \Delta W) x = W_0 x + \Delta W x = W_0 x + B(CD)A x
\end{aligned}
\end{equation}
where \(C \in \mathbb{R}^{r \times r}\), \(D \in \mathbb{R}^{r \times r}\), as illustrated in Figure \ref{P1} (C).

\textbf{Structure 4 DTSM (Dual Transformation Superposition Matrix)}: This structure utilizes the additive operation of matrices \(C\) and \(D\) to form a superposition transformation matrix, facilitating complex interactions between model parameters. This structure allows for detailed scaling and modulation of features, thereby providing enhanced expressiveness and flexibility to the model to address complex task requirements. Its forward propagation mathematical expression in the model is shown as \ref{eq7}:
\begin{equation}\label{eq7}
\begin{aligned}
h = (W_0 + \Delta W) x = W_0 x + \Delta W x = W_0 x + B(C+D)A x
\end{aligned}
\end{equation}
where \(C \in \mathbb{R}^{r \times r}\), \(D \in \mathbb{R}^{r \times r}\). This approach allows for more complex interactions between \(A\) and \(B\), enhancing the model's expressiveness and flexibility, as illustrated in Figure \ref{P1} (D).

The structural schematic of the MTLoRA fine-tuning method, as shown in Figure \ref{P1}.

The MTLoRA method can be applied not only in Dense Layers based on the Transformer architecture model but also in any neural network structure with Dense Layers.

\section{Experiments}

This experiment aims to validate the effectiveness of the MTLoRA method in natural language processing tasks. Specifically, the experimental design includes a total of eleven tasks in two major categories: NLU and NLG. GPT-2 Medium~\citep{radford2019language} and RoBERTa-base~\citep{liu2019roberta} are selected as the base pre-trained models, upon which fine-tuning is performed using MTLoRA and LoRA techniques. To comprehensively evaluate the performance of the MTLoRA method, the experimental results of MTLoRA will be compared and analyzed against existing fine-tuning methods, including Full Fine-Tuning, Adaptive Tuning, Bitfit, and LoRA.

\subsection{Baseline}
This study selected recently representative fine-tuning methods for comparative validation to evaluate the performance of each method on specific tasks. Below is a detailed description of each fine-tuning method:

\textbf{Full Fine-Tuning} involves updating all trainable parameters during the fine-tuning process of downstream tasks. Full fine-tuning typically achieves better performance but requires more GPU computational resources, larger storage space, and longer training time~\citep{qiu2020pre,raffel2020exploring}.

\textbf{Adapter Tuning} is an addition-based fine-tuning method that introduces extra trainable extension structures within the blocks of a Large Pre-trained Language Model (LPLM). During the fine-tuning process, the original parameters of the LPLM are frozen, with only the trainable parameters in the newly added extension structures being updated. Compared to full-parameter fine-tuning, adapter tuning significantly reduces the number of trainable parameters in the model and can achieve similar or even superior performance~\citep{houlsby2019parameter,rebuffi2017learning,he2022towards}.

\textbf{BitFit} is a method that specifies the updating of Bias parameters in LPLM while freezing the rest of the parameters. As described by~\cite{zaken2021bitfit}, BitFit significantly reduces the amount of parameter updates during the fine-tuning process and is able to maintain good performance.

In the replication experiments, the \textbf{LoRA} fine-tuning method maintained all hyperparameter configurations consistent with those described by~\cite{hu2022lora}.

\textbf{Ada-LoRA} is typically applied to all linear layer parameter matrices in LPLM~\citep{zhang2023adaptive}, but for effective comparison, we utilized the RoBERTa-base pre-trained model and applied the Ada-LoRA method to the query/value projection matrices $W_q$ and $W_v$, with experimental data derived from the Delta-LoRA~\citep{zi2023delta} experiments.

\textbf{Delta-LoRA} uses the hyperparameter settings of LoRA as its baseline configuration to facilitate direct performance comparisons across various models~\citep{zi2023delta}. In natural language understanding (NLU) tasks, Delta-LoRA reduces the input sequence length from 512 to 256 to decrease GPU memory demand and speed up the training process. Moreover, the batch size for different tasks is increased, for instance, from 16 to 120, with the update rate ($\lambda$) set to 0.5. Apart from these changes, the remaining hyperparameter configurations are consistent with LoRA. For natural language generation (NLG) tasks, the update rate ($\lambda$) is adjusted to 2. The Delta-LoRA method is applied to the query/value projection matrices $W_q$ and $W_v$ in LPLM.

\subsection{Natural Language Understanding Tasks}

\subsubsection{Base Model and Dataset}

In this experiment, we employed RoBERTa-base as the foundational pre-trained model, which was introduced by ~\cite{liu2019roberta} through the Facebook AI Research Lab. RoBERTa is an advancement over the BERT pre-training model, designed to enhance the model's performance on downstream tasks. RoBERTa implemented several improvements over BERT, including: (1) training the model with larger batch sizes, over longer periods, and with more data; (2) removing the Next Sentence Prediction (NSP) from the optimization objectives; (3) employing dynamic masking techniques during training; and (4) utilizing longer sentences in the training corpus. Due to its superior performance and moderate parameter size, the RoBERTa pre-trained model is widely utilized in performance testing for various downstream tasks based on pre-trained models.

To validate the effectiveness of the MTLoRA fine-tuning method in natural language understanding tasks, we conducted empirical experiments using the General Language Understanding Evaluation (GLUE) benchmark~\citep{wang2018glue}, which includes eight tasks, specifically:

\textbf{CoLA (Corpus of Linguistic Acceptability)}: Created by~\cite{warstadt2019neural}, this dataset comprises a series of English sentences sourced from books and journals on linguistic theory, aimed at assessing the grammatical acceptability of English sentences. The classification results are categorized as "acceptable" and "unacceptable." CoLA contains 8,551 training samples, 1,043 validation samples, and 1,063 test samples. The Matthews correlation coefficient~\citep{matthews1975comparison} is used as the performance evaluation metric, suitable for imbalanced binary classification tasks.

\textbf{MNLI (Multi-Genre Natural Language Inference)}: Introduced by~\cite{williams2017broad}, each sample consists of a premise sentence and a hypothesis sentence, with the primary task being to judge the textual entailment relationship between these two sentences. MNLI sentences cover ten genres, including transcripts of speeches, fiction, news reports, and government reports, requiring the model to handle texts of various styles, types, and domains. MNLI includes 392,702 training samples, 19,647 validation samples, and 19,647 test samples.

\textbf{MRPC (Microsoft Research Paraphrase Corpus)}: Proposed by~\cite{dolan2005automatically}, the dataset contains sentence pairs automatically extracted from news sources, primarily to determine whether two sentences have a paraphrasing relationship. The distribution of these two categories is imbalanced, with about 68\% being positive samples, where the semantic similarity of sentence pairs is manually annotated. MRPC includes 3,668 training samples, 408 validation samples, and 1,725 test samples.

\textbf{QNLI (Question Natural Language Inference)}: Provided by~\cite{wang2018glue}, samples are derived from the Stanford Question Answering Dataset~\citep{rajpurkar2016squad} and converted into a classification task. The main task is to determine whether a given statement sentence contains the answer to a given question sentence, with sentences collected from Wikipedia and questions manually written. QNLI contains 104,743 training sample pairs, 5,463 validation sample pairs, and 5,463 test sample pairs.

\textbf{QQP (Quora Question Pairs)}: Offered by the question-and-answer website Quora, the main task is to determine whether two question sentences are semantically similar. The distribution of these two categories is imbalanced, with about 63\% being negative samples. QQP includes 363,846 training samples, 40,430 validation samples, and 390,965 test samples~\citep{wang2018glue}.

\textbf{RTE (Recognizing Textual Entailment)}: Combines results published by multiple research organizations~\citep{dagan2005pascal,haim2006second,giampiccolo2007third,bentivogli2009fifth} between 2005 and 2011 during the RTE challenges. The sample data, composed of Wikipedia and news data, has a primary task similar to MNLI but on a smaller scale, including 2,490 training samples, 277 validation samples, and 3,000 test samples.

\textbf{SST-2 (Stanford Sentiment Treebank)}: Introduced by~\cite{socher2013recursive}, the main task is to determine the sentiment orientation of a given sentence, whether positive or negative. The sample data mainly comes from movie review data, with reviews manually annotated for sentiment classification. SST-2 includes 67,349 training samples, 872 validation samples, and 1,821 test samples.

\textbf{STS-B (Semantic Textual Similarity Benchmark)}: Proposed by~\cite{cer2017semeval}, the main task is to calculate the semantic similarity score between two sentences. Samples are extracted from sentence pairs collected from videos, image captions, news headlines, etc., with sentence pair similarity manually rated on a scale from 1 to 5. Model performance is evaluated using the Pearson and Spearman correlation coefficients. STS-B includes 5,749 training samples, 1,500 validation samples, and 1,379 test samples.

\subsubsection{Experimental Details}
\label{NLU_Detail}

This experiment was conducted on the basis of the RoBERTa-base as the pre-trained model, applying both MTLoRA and LoRA methods across eight tasks on the GLUE benchmark. The following settings were employed during the training and testing process:
\begin{itemize}
\item AdamW was chosen as the optimizer for the model, along with a linear learning rate decay strategy;
\item The GPU processor used was NVIDIA A100-PCIE-40GB;
\item The hyperparameter configuration was consistent with the settings by~\cite{hu2022lora}, including learning rate, maximum sequence length (MaxSeqLength), rank (\(r\)) size, batch size (BatchSize), and epochs. Detailed hyperparameter settings can be found in Appendix \ref{RoBERTa_super_para};
\item For each task, the reported experimental results are based on the median and standard deviation obtained from training and testing with three different random seeds;
\item The MTLoRA method was applied to the query/value projection matrices (\(W_q\), \(W_v\)), intermediate layer projection matrix (\(W_m\)), and output layer projection matrix (\(W_o\)), while the LoRA method was applied to the query/value projection matrices (\(W_q\), \(W_v\)).
\end{itemize}

\subsubsection{Experimental Results}
In this experiment, we explored the impact of different $T$ transformation matrix structures of the MTLoRA method on natural language understanding tasks. The test experiment results of the LoRA and MTLoRA methods based on the RoBERTa-base model across eight tasks of the GLUE benchmark are shown in Table \ref{Table1}:

\begin{table*}[htbp]
    \centering
        \resizebox{\textwidth}{!}{
    \begin{tabular}{m{3.4cm}|cccccccc|c}
        \hline \rule{0pt}{2.5ex}\textbf{Model\&Method (RoBERTa-base)} & \textbf{MNLI} & \textbf{SST-2} & \textbf{CoLA} & \textbf{QQP} & \textbf{QNLI} & \textbf{RTE} & \textbf{MRPC} & \textbf{STS-B} & \textbf{Avg.} \\
        \hline
        \rule{0pt}{2.5ex}Full Fine-tuning & $87.6$ & $94.8$ & $63.6$ & $91.9$ & $92.8$ & $78.7$ & $90.2$ & $91.2$ & $86.35$ \\
        BitFit & $84.7$ & $93.7$ & $62$ & $84$ & $91.8$ & $81.5$ & $92.7$ & $90.8$ & $85.15$ \\
        Adapter & $87.3_{\pm .1}$ & $94.7_{\pm .3}$ & $62.6_{\pm .9}$ & $90.6_{\pm .0}$ & $93.0_{\pm .2}$ & $75.9_{\pm 2.2}$ & $88.4_{\pm .1}$ & $90.3_{\pm .1}$ & $85.4$ \\
        Ada-LoRA(best) & $87.34$ & $94.49$ & $61.64$ & $90.14$ & $93.08$ & $85.19$ & $90.19$ & $91.16$ & $86.65$ \\
        Delta-LoRA(best) & $87.5$ & $95.06$ & $63.82$ & $90.87$ & $93.09$ & $87$ & $90.19$ & $91.57$ & $87.38$ \\
        LoRA(best) & $87.49$ & $94.61$ & $61.82$ & $90.41$ & $92.9$ & $85.2$ & $88.97$ & $91.43$ & $86.6$ \\
        LoRA & $87.37_{\pm .1}$ & $94.38_{\pm .1}$ & $61.57_{\pm .8}$ & $90.37_{\pm .0}$ & $92.6_{\pm .1}$ & $84.12_{\pm .7}$ & $88.24_{\pm .4}$ & $91.27_{\pm .1}$ & $86.24$ \\
        MTLoRA(best) & $\mathbf{87.54}$ & $\mathbf{95.41}$ & $\mathbf{65.29}$ & $\mathbf{91.4}$ & $\mathbf{92.75}$ & $\mathbf{87.73}$ & $\mathbf{90.93}$ & $\mathbf{91.39}$ & $\mathbf{87.8}$ \\
        MTLoRA & $\mathbf{87.49}_{\pm .1}$ & $\mathbf{94.5}_{\pm .1}$ & $\mathbf{63.11}_{\pm .1}$ & $\mathbf{91.25}_{\pm .0}$ & $\mathbf{92.6}_{\pm .2}$ & $\mathbf{87.73}_{\pm .8}$ & $\mathbf{90.69}_{\pm .3}$ & $\mathbf{91.37}_{\pm .0}$ & $\mathbf{87.34}$ \\
        \hline
        \rule{0pt}{2.5ex}MTLoRA-SHIM(best) & $\mathbf{87.54}$ & $\mathbf{95.41}$ & $\mathbf{65.29}$ & $91.33$ & $92.62$ & $\mathbf{87.73}$ & $90.44$ & $\mathbf{91.39}$ & $\mathbf{87.71}$ \\
        MTLoRA-SHIM & $\mathbf{87.49}_{\pm .1}$ & $\mathbf{94.5}_{\pm .1}$ & $62.86_{\pm 1.4}$ & $91.24_{\pm .0}$ & $\mathbf{92.6}_{\pm .2}$ & $\mathbf{87.73}_{\pm .8}$ & $89.71_{\pm .4}$ & $\mathbf{91.37}_{\pm .0}$ & $\mathbf{87.18}$ \\
        MTLoRA-ICFM(best) & $87.43$ & $94.84$ & $62.83$ & $91.19$ & $92.37$ & $85.56$ & $90.2$ & $\mathbf{91.34}$ & $86.97$ \\
        MTLoRA-ICFM & $87.13_{\pm .1}$ & $94.38_{\pm .3}$ & $62.57_{\pm 1.0}$ & $91.08_{\pm .0}$ & $92.35_{\pm .0}$ & $85.2_{\pm .6}$ & $89.46_{\pm .6}$ & $90.95_{\pm .2}$ & $86.64$ \\
        MTLoRA-CTCM(best) & $87.33$ & $\textbf{95.41}$ & $63.85$ & $91.11$ & $92.39$ & $\mathbf{87}$ & 89.95 & 91.38 & 87.3 \\
        MTLoRA-CTCM & $87.32_{\pm .1}$ & $94.38_{\pm .5}$ & $62.57_{\pm .6}$ & $91.0_{\pm .0}$ & $92.15_{\pm .1}$ & $85.92_{\pm .8}$ & $89.71_{\pm .5}$ & $91.34_{\pm .0}$ & $86.79$ \\
        MTLoRA-DTSM(best) & $87.07$ & $94.84$ & $63.33$ & $\mathbf{91.4}$ & $\mathbf{92.75}$ & $85.56$ & $\mathbf{90.93}$ & $91.2$ & $87.13$ \\
        MTLORA-DTSM & $86.98_{\pm .0}$ & $94.27_{\pm .4}$ & $\mathbf{63.11}_{\pm .1}$ & $\mathbf{91.25}_{\pm .0}$ & $92.44_{\pm .1}$ & $85.2_{\pm .2}$ & $\mathbf{90.69}_{\pm .3}$ & $91.12_{\pm .0}$ & $86.88$ \\
        \hline
    \end{tabular}}
	\caption{The table presents the experimental results of fine-tuning methods such as MTLoRA and LoRA based on the RoBERTa-base model across the eight datasets of the GLUE benchmark. Among these, MNLI and QNLI are measured by accuracy, CoLA by Matthew's correlation, and STS-B by Pearson correlation. For all these metrics, higher values represent better performance.}
	\label{Table1}
\end{table*}

We use the results of LoRA~\citep{hu2022lora} as the primary benchmark for comparison. From the results in Table 1, it can be seen that MTLoRA achieves an average performance improvement of about 1\% across all tasks and controls the standard deviation fluctuations well. Specifically:

The SHIM structure exhibits enhanced performance and stability across a majority of tasks, with observed performance gains of roughly 1.29\% ($\sigma$=1.4\%) on CoLA, 0.87\% ($\sigma$=0.0\%) on QQP, and 3.61\% ($\sigma$=0.8\%) on RTE tasks. This performance indicates its effectiveness and wide-ranging utility in a variety of general tasks.

In contrast, the ICFM structure shows robust performance across numerous tasks, particularly excelling in tasks related to semantic similarity and entailment when there is a lack of comprehensive corpus data. For example, it achieves an average performance boost of approximately 1.22\% ($\sigma$=0.6\%) on the MRPC task and 1.08\% ($\sigma$=0.6\%) on the RTE task.

Meanwhile, the CTCM structure delivers consistent performance across a spectrum of tasks, standing out particularly in inference tasks with extensive corpus data. It demonstrates a performance increase of 1.0\% ($\sigma$=0.6\%) on the CoLA task, indicating a noteworthy reduction in standard deviation when compared to the SHIM structure.

Lastly, the DTSM structure continues to show steady performance in various tasks, particularly highlighting an improvement of around 0.88\% ($\sigma$=0.0\%) on the QQP task when there is ample corpus data, and a mean performance enhancement of about 1.54\% ($\sigma$=0.1\%) on the CoLA task. This significant decrease in standard deviation underscores the DTSM structure's efficiency in managing semantic similarity tasks with abundant corpora.

\begin{figure*}[htb]
  \centering
  \includegraphics [width=14cm]{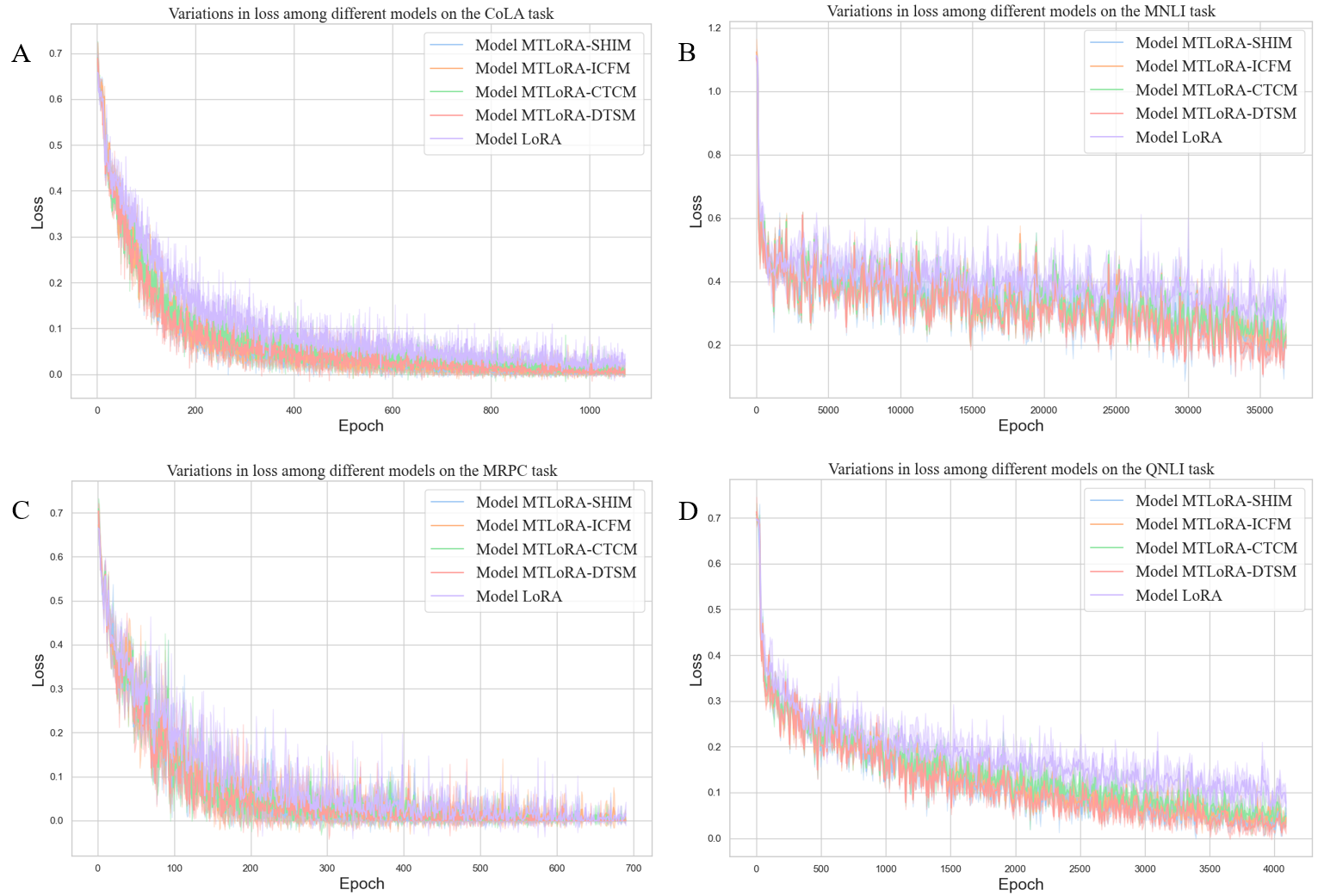}
  \caption{The changes in loss for different transformation matrix structures of the MTLoRA model and the LoRA model on the CoLA, MNLI, MRPC, and QNLI tasks.}
  \label{P3}
\end{figure*}

\begin{figure*}[htb]
  \centering
  \includegraphics [width=14cm]{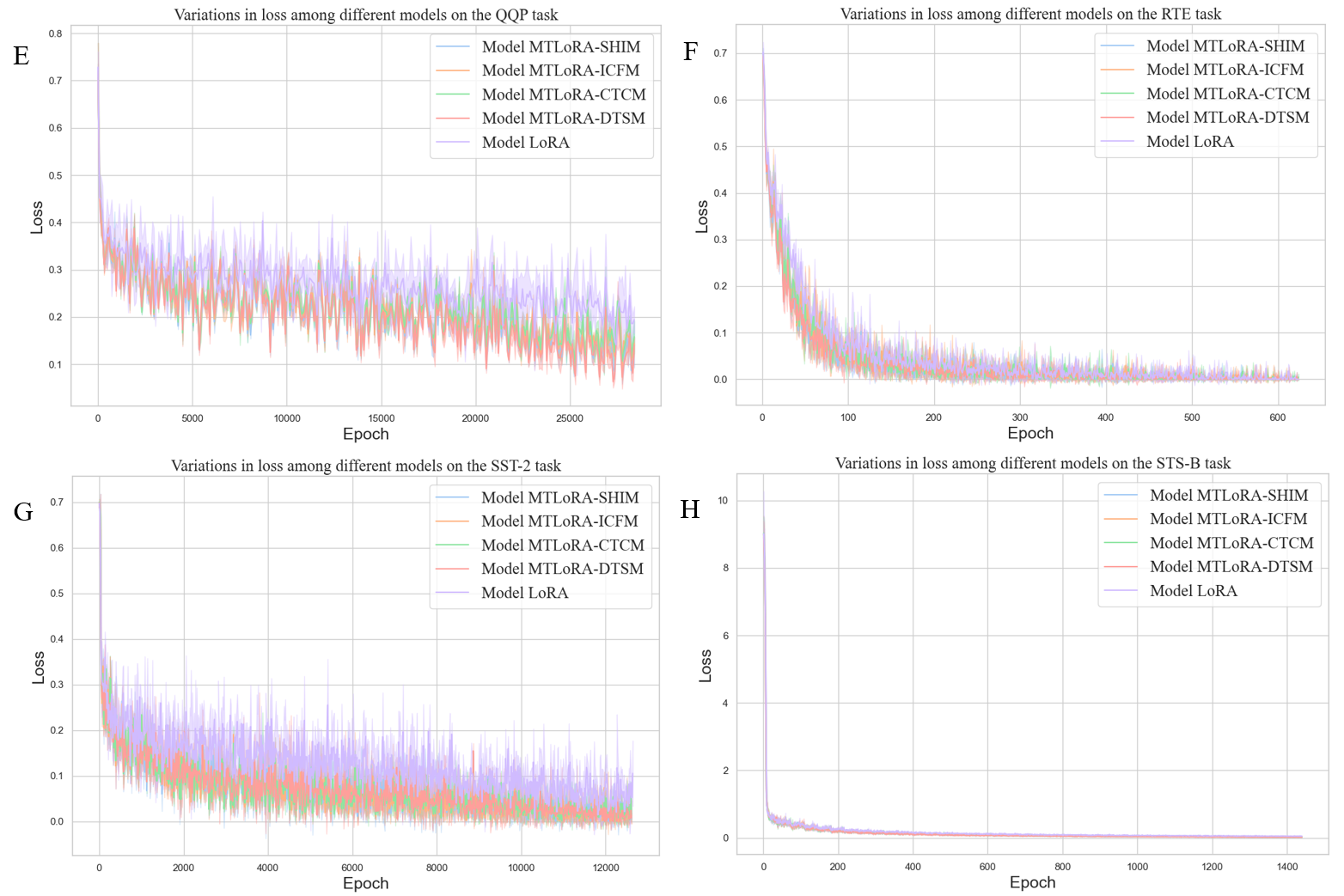}
  \caption{The changes in loss for different transformation matrix structures of the MTLoRA model and the LoRA model on the QQP, RTE, SST-2, and STS-B tasks.}
  \label{P4}
\end{figure*}

During the training process, the changes in loss for different transformation matrix structures of the MTLoRA model and the LoRA model across the eight tasks of the GLUE benchmark are illustrated in Figures \ref{P3} and \ref{P4}. From the figures, it can be observed that compared to LoRA, the MTLoRA method exhibits lower loss, faster convergence speed, and lower standard deviation of fluctuations, indicating that MTLoRA indeed can enhance model performance on multiple tasks.

Overall, the different transformation matrix structures in the MTLoRA method may demonstrate superior performance on various tasks. Therefore, in practical applications, an appropriate transformation matrix structure should be selected.

\subsection{Natural Language Generation Tasks}

\subsubsection{Base Model and Datasets}
In this experiment, we employ the GPT-2 Medium model released by OpenAI in 2019 as the pre-trained base model. This model is widely used in various text generation tasks due to its strong natural language generation capabilities~\citep{radford2019language}. We conducted experiments on three broadly recognized datasets: the E2E NLG Challenge, WebNLG, and DART.

The \textbf{E2E NLG Challenge} dataset was introduced by~\cite{novikova2017e2e}, primarily aiming to advance the research of end-to-end, data-to-text natural language generation systems. This dataset poses multiple challenges for the performance of natural language generation systems, including a large number of samples, complex syntactic structure variations, rich vocabulary, and diverse sentence structures. The sample content pertains to the restaurant domain, requiring the system to generate detailed and fluent sentences based on given structured information. The dataset contains approximately 42,000 training samples, 4,600 validation samples, and 4,600 test samples.

The \textbf{WebNLG} dataset was introduced by~\cite{gardent2017webnlg}, and its primary task is to convert structured RDF (Resource Description Framework) triple data into fluent natural language text, facilitating the training and performance evaluation of generation systems. WebNLG thoroughly examines the system's capabilities in micro-planning, which includes several sub-tasks such as sentence segmentation, lexicalization, referring expression generation, information aggregation, and surface realization. The dataset focuses on 14 categories within the DBpedia domain and specifically notes that 5 of these categories are not included in the training set but only appear in the test set, to assess the model's generalization ability on unseen data. The dataset comprises approximately 22K samples.

The \textbf{DART} dataset was introduced by~\cite{nan2020dart}, with the primary task of converting structured triple data into high-quality natural language text for training and performance evaluation of generation systems. DART is characterized by its large scale, broad domains, and a robust structured data ontology semantic representation framework. Unlike the E2E and WebNLG datasets, which utilize a flat slot-value ontology representation structure, DART employs a unique tree-shaped ontology semantic representation framework. This framework more effectively encodes the rich semantic dependencies between ontologies in structured data.

\subsubsection{Experimental Details}

This experiment conducts tests on the E2E NLG Challenge, WebNLG, and DART tasks using the MTLoRA and LoRA methods, with the GPT-2 Medium model serving as the pre-trained base. The following settings were adopted during the training and testing phases:
\begin{itemize}
\item The AdamW optimizer was chosen for the model, along with a linear learning rate decay strategy.
\item The GPU used was the NVIDIA A100-PCIE-40GB.
\item The hyperparameter configuration was consistent with the settings of~\cite{hu2022lora}, including learning rate, BeamSize, rank (\(r\)) size, batch size (BatchSize), training cycles (Epoch), etc. Detailed hyperparameter settings can be found in Appendix \ref{GPT-2_super_para}.
\item For each task, the reported experimental results are the averages and standard deviations based on three sets of training and testing with different random seeds.
\item The MTLoRA method was applied to the weight matrices in the query/key/value projection matrices (\(W_q\), \(W_k\), \(W_v\)) and the MLP layer (\(W_f\)), while the LoRA method was applied to the query/value projection matrices (\(W_q\), \(W_v\)).
\item Since the query/key/value projection matrices (\(W_q\), \(W_k\), \(W_v\)) of the GPT-2 Medium model are generated in a merged manner, this experiment utilized the MergedLinear stepwise convolution structure to address this situation. MergedLinear implements stepwise convolution through a transformation matrix, $T$. Specifically, first, the parameter matrix $A$ uses the transformation matrix $T$ as a convolution kernel to perform joint convolution across the $K$, $Q$, $V$ channels, obtaining the transformed matrix $A^{\prime}$ after the first convolution step. Matrix $A^{\prime}$ can integrate the associated information within the $KQV$ structure, beneficial for enhancing the model's representational capability. Subsequently, matrix $A^{\prime}$ undergoes independent convolution on the $K$, $Q$, $V$ channels using the parameter matrix $B$ as a convolution kernel, resulting in the parameter increment matrix $\Delta W$ after the second convolution step, where $r$ << $k$ or $d$. The structural diagram is illustrated in Figure \ref{P2}.

\begin{figure*}[htb]
  \centering
  \includegraphics [width=12cm]{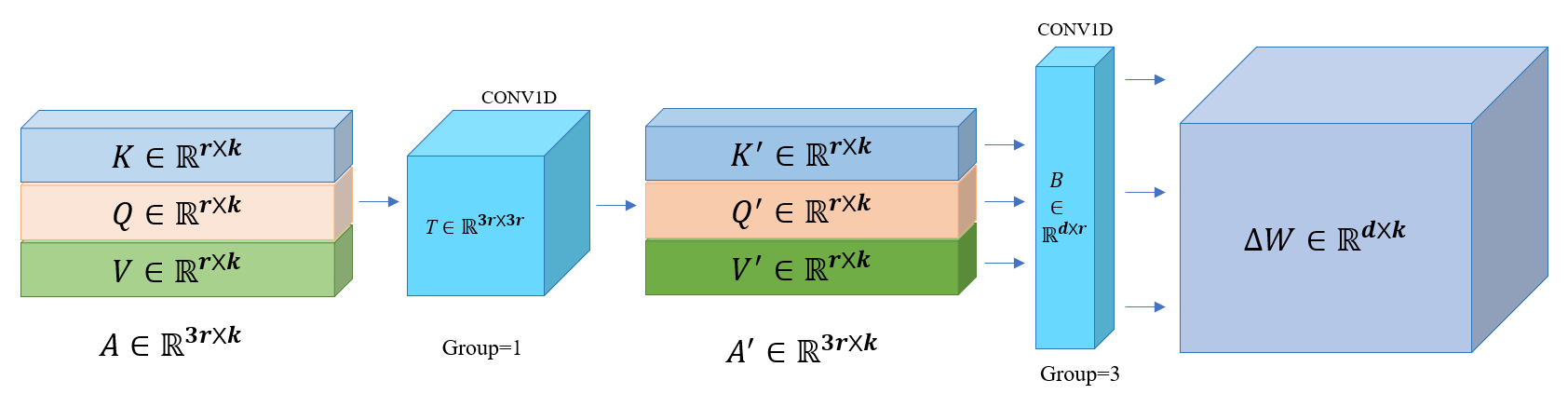}
  \caption{Stepwise convolution structure diagram of the MTLoRA fine-tuning method.}
  \label{P2}
\end{figure*}

\end{itemize}

\subsubsection{Experimental results}

In this experiment, we explored the impact of different $T$ transformation matrix structures of the MTLoRA method on natural language generation tasks. The test experiment results of the LoRA and MTLoRA methods, based on the GPT-2 Medium pre-trained model on the E2E, DART, and WebNLG datasets, are presented in Table \ref{Table2}:

\begin{table*}[htbp]
    \centering
    \small
    \begin{tabular}{l|ccc}
        \hline \rule{0pt}{2.5ex}\textbf{Model\&Method (GPT-2 Medium)} & \textbf{E2E(BLEU)} & \textbf{DART(BLEU)} & \textbf{WebNLG(BLEU-A)} \\
        \hline
        \rule{0pt}{2.5ex}Full Fine-Tune & $68.2$ & $46.2$ & $46.5$ \\
        Adapter & $66.3$ & $42.4$ & $50.2$ \\
        Prefix & $69.7$ & $46.4$ & $55.1$ \\
        Ada-LoRA (best) & $68.16$ & $/$ & $50.82$ \\
        Delta-LoRA(best) & $70.84$ & $/$ & $55.96$ \\
        LoRA & $70.4_{\pm .1}$ & $47.1_{\pm .2}$ & $55.3_{\pm .2}$ \\
        MTLoRA(best) & $70.23$ & $48.31$ & $55.98$ \\
        MTLoRA & $70.02_{\pm .2}$ & $\mathbf{48.05}_{\pm .1}$ & $\mathbf{55.86}_{\pm .1}$ \\
        \hline
        \rule{0pt}{2.5ex}MTLoRA-SHIM(best) & $70.1$ & $48$ & $55.98$ \\
        MTLoRA-SHIM & $69.58_{\pm .3}$ & $47.93_{\pm .0}$ & $\mathbf{55.86}_{\pm .1}$ \\
        MTLoRA-ICFM(best) & $70.23$ & $47.9$ & $55.98$ \\
        MTLoRA-ICFM & $70.02_{\pm .2}$ & $47.75_{\pm .1}$ & $55.52_{\pm .5}$ \\
        MTLoRA-CTCM(best) & $69.88$ & $48$ & $55.79$ \\
        MTLoRA-CTCM & $68.88_{\pm .8}$ & $47.98_{\pm .0}$ & $55.57_{\pm .2}$ \\
        MTLoRA-DTSM(best) & $69.43$ & $48.31$ & $55.86$ \\
        MTLoRA-DTSM & $68.61_{\pm .5}$ & $\mathbf{48.05}_{\pm .1}$ & $55.61_{\pm .3}$ \\
        \hline
    \end{tabular}
	\caption{The table displays the experimental results of fine-tuning methods such as MTLoRA and LoRA based on the GPT-2 Medium model on the E2E, DART, and WebNLG datasets. All tasks use the BLEU metric for evaluation, with higher values indicating better performance.}
	\label{Table2}
\end{table*}

We use the experimental results of LoRA~\citep{hu2022lora} as the primary benchmark for comparison. From the results in Table \ref{Table2}, it is evident that the four transformation matrix structures of the MTLoRA fine-tuning method effectively enhance the model's performance on the DART and WebNLG tasks. Specifically:

The incorporation of the SHIM structure led to an enhancement in the model's efficacy on the DART and WebNLG tasks, with an improvement of approximately 0.83\% ($\sigma$=0.0\%) and 0.56\% ($\sigma$=0.1\%), respectively. Similarly, the implementation of the ICFM structure resulted in a performance increment of roughly 0.65\% ($\sigma$=0.1\%) for DART and 0.22\% ($\sigma$=0.5\%) for WebNLG. The application of the CTCM structure contributed to a performance uplift on the DART and WebNLG tasks by approximately 0.88\% ($\sigma$=0.0\%) and 0.27\% ($\sigma$=0.2\%), respectively. Lastly, the adoption of the DTSM structure facilitated a performance boost on the DART and WebNLG tasks by an estimated 0.95\% ($\sigma$=0.1\%) and 0.31\% ($\sigma$=0.3\%), respectively.

The experimental results of MTLoRA on natural language generation tasks once again demonstrate that different transformation matrix structures of the MTLoRA fine-tuning method may achieve optimal performance on different tasks. Particularly, when the query/key/value projection matrices ($W_q$, $W_k$, $W_v$) of the pre-trained model are generated jointly, the stepwise convolutional transformation design of MTLoRA can effectively integrate and share the associated information of these three components, better supporting downstream tasks.

\subsection{Experimental Analysis}

To further validate the effectiveness of the MTLoRA method, we adapted LoRA to the query/value projection matrices ($W_q$, $W_v$), the intermediate layer projection matrix ($W_m$), and the output layer projection matrix ($W_o$) of the RoBERTa-base model. The other experimental details remain consistent with the description in section \ref{NLU_Detail}. The experimental results on the GLUE benchmark for natural language understanding are shown in Table \ref{Table3}:

\begin{table*}[htbp]
    \centering
        \resizebox{\textwidth}{!}{
    \begin{tabular}{m{3cm}|cccccccc|c}
        \hline \rule{0pt}{2.5ex}\textbf{Model\&Method (roberta-base)} & \textbf{MNLI} & \textbf{SST-2} & \textbf{CoLA} & \textbf{QQP} & \textbf{QNLI} & \textbf{RTE} & \textbf{MRPC} & \textbf{STS-B} & \textbf{Avg.} \\
        \hline 
        \rule{0pt}{2.5ex}LoRA*(best) & $87.49$ & $94.61$ & $61.82$ & $90.4$1 & $92.9$ & $85.2$ & $88.97$ & $91.43$ & $86.6$ \\
        LoRA* & $87.33_{\pm .1}$ & $94.38_{\pm .1}$ & $61.07_{\pm .8}$ & $90.37_{\pm .0}$ & $92.68_{\pm .1}$ & $84.23_{\pm .7}$ & $88.39_{\pm .4}$ & $91.23_{\pm .1}$ & $86.21$ \\
        LoRA & $87.3_{\pm .2}$ & $94.27_{\pm .3}$ & $62.83_{\pm 1.1}$ & $\mathbf{88.35}_{\pm \mathbf{12.5}}$ & $92.92_{\pm .1}$ & $\mathbf{84.84}_{\pm \mathbf{2.0}}$ & $88.97_{\pm .3}$ & $91.29_{\pm 0.0}$ & $86.34$ \\
        \hline 
        \rule{0pt}{2.5ex}LoRA(best) & $87.44$ & $94.95$ & $63.33$ & $91.05$ & $93.08$ & $86.64$ & $89.46$ & $91.31$ & $87.15$ \\
        LoRA(least) & $86.92$ & $94.04$ & $60.58$ & $\mathbf{63.18}$ & $92.7$ & $\mathbf{81.59}$ & $88.73$ & $91.21$ & $82.36$ \\
        \hline
    \end{tabular}}
	\caption{The table shows the experimental results of adapting LoRA to the query/value projection matrices ($W_q$, $W_v$), the intermediate layer projection matrix ($W_m$), and the output layer projection matrix ($W_o$) of the RoBERTa-base model on the GLUE benchmark for natural language understanding. $*$ indicates that the data are sourced from Table \ref{Table1}.}
	\label{Table3}
\end{table*}

From the experimental results in Table \ref{Table3}, LoRA shows a slight improvement in performance on the QNLI and MRPC tasks of the GLUE benchmark. However, there is a significant fluctuation in performance on the QQP task, with a standard deviation as high as 12.5\%, and a 2\% performance fluctuation on the RTE task. This indicates that merely increasing the amount of trainable parameters does not always lead to performance improvements and can sometimes result in performance degradation.

\textbf{Performance of LoRA on E2E Tasks.} To verify whether the MTLoRA method actually reduces the model's performance on the E2E task in NLG, we re-executed the performance test of LoRA on the E2E task using the same computational environment as MTLoRA. The experimental results show that LoRA achieves an average BLEU score of 69.52\% on the E2E task, with a standard deviation of 0.5\%. Its average BLEU score is comparable to that of MTLoRA, but with an increased standard deviation. This indicates that the experimental computational environment also has an impact on the results, suggesting that MTLoRA does not actually reduce the model's performance on the E2E task in NLG.

\textbf{Sensitivity Analysis of Rank ($r$).} This study employs the MTLoRA and LoRA methods based on RoBERTa-base, conducting experiments on the COLA task with varying sizes of rank ($r$) to assess the specific impacts of rank ($r$) on model efficacy and stability. All experiments were carried out with the same random seed, and the specific settings of the experiments are consistent with those described in Section \ref{NLU_Detail}, with the corresponding results displayed in Figure \ref{P5}:

\begin{figure*}[htb]
  \centering
  \includegraphics [width=12cm]{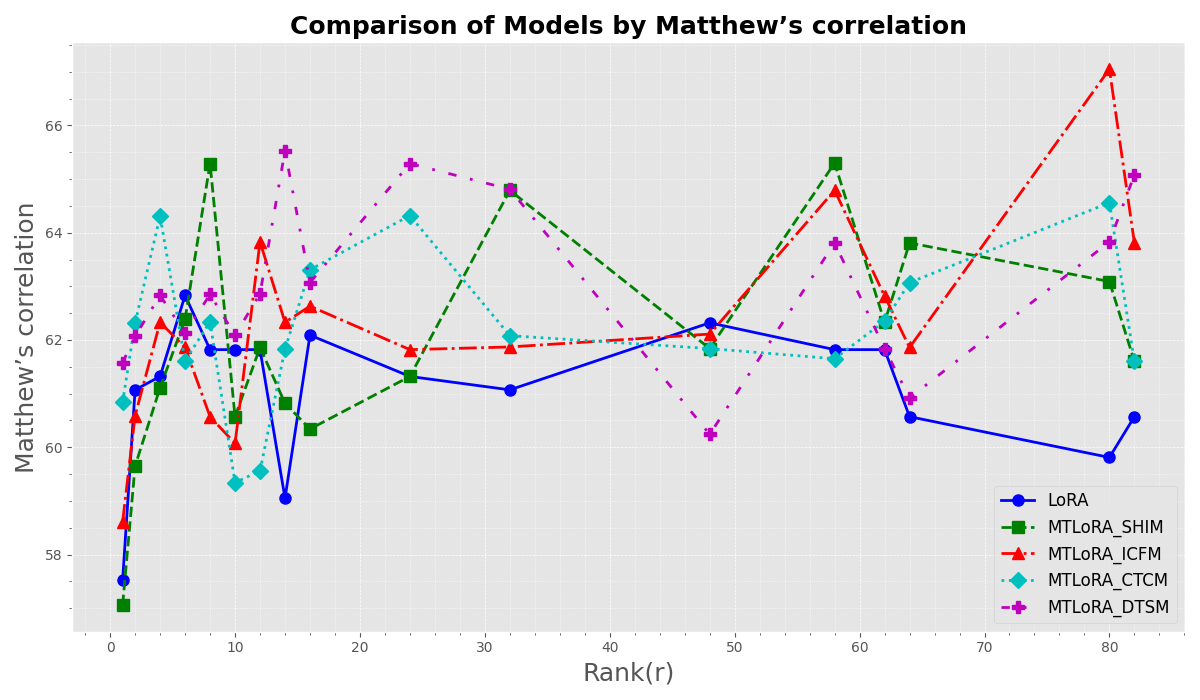}
  \caption{The MTLoRA and LoRA methods, based on the RoBERTa-base model, were experimentally evaluated on the COLA task using different rank ($r$) sizes. A higher Matthew's correlation value indicates better model performance.}
  \label{P5}
\end{figure*}

Several observations can be made from the experimental results in Figure \ref{P5}: Firstly, the MTLoRA method continues to exhibit good performance and stability on the COLA task across different rank sizes without evident overfitting. Particularly, the ICFM structure in MTLoRA demonstrates exceptional performance and stability, indicating that the ICFM structure can effectively capture the task's key features, thereby efficiently preventing overfitting. Secondly, compared to the traditional LoRA, MTLoRA can achieve performance improvements with larger ranks while adding fewer parameters. Lastly, the effectiveness of MTLoRA is not solely due to the increased parameters at the same rank, as the performance curve within the examined range suggests that MTLoRA can outperform some higher-ranked LoRAs even at lower ranks.

\section{Conclusion and Future Work}

In this paper, we explored fine-tuning techniques based on LPLMs, which efficiently adjust a small number of parameters to significantly reduce the computational power and space requirements of the model while maintaining its performance on downstream tasks. Among these, the reparameterization-based fine-tuning technique represented by LoRA stands out, being widely applied to various task scenarios. However, LoRA and its improved methods still have shortcomings in complex task adaptability, performance, stability, and algorithmic complexity.

To mitigate these issues, we proposed MTLoRA. This method applies linear transformations to task-specific parameter matrices through the $T$ transformation matrix, altering their spatial geometric structure to generate new matrix feature patterns. This mimics the fundamental impact of different geometric structural feature patterns in the brain on functions, thereby alleviating the aforementioned problems. Through a series of experimental validations, the MTLoRA method has improved performance in multiple task scenarios compared to the LoRA method while also reducing standard deviation. Notably, the MTLoRA method not only enhances performance but also maintains the simplicity of the algorithm, without increasing inference stage latency or reducing the length of the input sequence. Furthermore, the MTLoRA method is applicable not only to the Transformer architecture but also to other neural network structures.

Future research will focus on continuously optimizing the MTLoRA method by integrating more in-depth principles of brain neuroscience. It will be applied in the field of safe alignment for large models, ensuring that the models not only perform exceptionally in specific tasks but also possess robust security capabilities. This will contribute to the sustainable development of large model applications. Moreover, considering the effectiveness of the SHIM, ICFM, CTCM, and DTSM transformation matrix structures, we plan to study more brain-inspired transformation matrix structures to further enhance the model's performance, reduce the standard deviation, and improve the model's adaptability in various downstream tasks. MTLoRA can also serve as a fundamental component to strengthen the model's capabilities in handling long contexts and parameter quantization.

\section{Acknowledgments}
This work was supported by the National Science and Technology Major Project (Grant No. 2022ZD0116202).

\bibliographystyle{frontiersinSCNS_ENG_HUMS}
\bibliography{MTLoRA}

\appendix

\section{The hyperparameter settings in the experiment}

\subsection{RoBERTa-base}
\label{RoBERTa_super_para}
We provide a detailed report on the hyperparameter configurations for the experiments conducted on the 8 datasets of the GLUE benchmark using the MTLoRA and LoRA methods, which are based on the RoBERTa-base pre-trained model, as shown in Table \ref{Table4}:

\begin{table*}[htbp]
    \centering
        \resizebox{\textwidth}{!}{
    \begin{tabular}{l|cccccccc}
        \hline \rule{0pt}{2.5ex}\textbf{Hyper-Parameter\&Dataset} & \textbf{MNLI} & \textbf{SST-2} & \textbf{CoLA} & \textbf{QQP} & \textbf{QNLI} & \textbf{RTE} & \textbf{MRPC} & \textbf{STS-B} \\
        \hline
        \rule{0pt}{2.5ex}Batch Size & 16 & 16 & 32 & 16 & 32 & 32 & 16 & 16 \\
        Number of Epochs & 30 & 60 & 80 & 25 & 25 & 80 & 30 & 40 \\
        Learning Rate & $5 \mathrm{E}-04$ & $5 \mathrm{E}-04$ & $4 \mathrm{E}-04$ & $5 \mathrm{E}-04$ & $4 \mathrm{E}-04$ & $5 \mathrm{E}-04$ & $4 \mathrm{E}-04$ & $4 \mathrm{E}-04$ \\
        Warmup Ratio & 0.06 & 0.06 & 0.06 & 0.06 & 0.06 & 0.06 & 0.06 & 0.06 \\
        Rank $\mathrm{r}$ & 8 & 8 & 8 & 8 & 8 & 8 & 8 & 8 \\
        Alpha $\alpha$ & 16 & 16 & 16 & 16 & 16 & 16 & 16 & 16 \\
        Max Sequence Length & 512 & 512 & 512 & 512 & 512 & 512 & 512 & 512 \\
        \hline
    \end{tabular}}
	\caption{The table provides the hyperparameter configurations for the experiments on the 8 datasets of the GLUE benchmark using the MTLoRA and LoRA methods, based on the RoBERTa-base model.}
	\label{Table4}
\end{table*}

\subsection{GPT-2 Medium}
\label{GPT-2_super_para}
We provide a detailed report on the hyperparameter configurations for the experiments conducted on the E2E, DART, and WebNLG datasets using the MTLoRA and LoRA methods, which are based on the GPT-2 Medium pre-trained model. Specifically, the configurations of hyperparameters during the training process and the inference process are shown in Table \ref{Table5} and Table \ref{Table6}, respectively:

\begin{table*}[htbp]
    \centering
    \small
    \begin{tabular}{l|ccc}
        \hline \rule{0pt}{2.5ex}\textbf{Hyper-Parameter\&Dataset} & \textbf{E2E} & \textbf{DART} & \textbf{WebNLG} \\
        \hline
        Batch Size & 8 & 8 & 8 \\
        Learning Rate & 0.0002 & 0.0002 & 0.0002 \\
        Number of Epochs & 5 & 5 & 5 \\
        Weight Decay & 0.01 & 0.0 & 0.01 \\
        Warmup Steps & 500 & 500 & 500 \\
        Dropout Prob & 0.1 & 0.0 & 0.1 \\
        Label Smooth & 0.1 & 0.0 & 0.1 \\
        Rank r & 4 & 4 & 4 \\
        Alpha $\alpha$ & 32 & 32 & 32 \\
        \hline
    \end{tabular}
	\caption{The table shows the hyperparameter configurations for the MTLoRA and LoRA methods based on the GPT-2 Medium model during the training process, in experiments on the E2E, DART, and WebNLG datasets.}
	\label{Table5}
\end{table*}

\begin{table*}[htbp]
    \centering
    \small
    \begin{tabular}{l|ccc}
        \hline \rule{0pt}{2.5ex}\textbf{Hyper-Parameter\&Dataset} & \textbf{E2E} & \textbf{DART} & \textbf{WebNLG} \\
        \hline
        Beam Size & 10 & 10 & 10 \\
        Length Penalty & 0.9 & 0.8 & 0.8 \\
        No Repeat Ngram Size & 4 & 4 & 4 \\
        \hline
    \end{tabular}
	\caption{The table displays the hyperparameter configurations for the MTLoRA and LoRA methods based on the GPT-2 Medium model during the inference process, in experiments on the E2E, DART, and WebNLG datasets.}
	\label{Table6}
\end{table*}

\end{document}